\documentclass{article}

\usepackage{microtype}
\usepackage{graphicx}
\usepackage{subfigure}
\usepackage{booktabs} 

\usepackage{hyperref}

\usepackage[accepted]{icml2023}

\usepackage{amsmath}
\usepackage{amssymb}
\usepackage{mathtools}
\usepackage{amsthm}

\usepackage[capitalize,noabbrev]{cleveref}

\theoremstyle{plain}

\theoremstyle{definition}

\theoremstyle{remark}

\usepackage[textsize=tiny]{todonotes}

\icmltitlerunning{Conformal Generative Modeling on Triangulated Surfaces}

\newcommand{\bb}[1]{\mathbb{ #1 }}
\newcommand{\C}{\bb{C}}
\newcommand{\R}{\bb{R}}

\begin{document}

\twocolumn[
\icmltitle{Conformal Generative Modeling on Triangulated Surfaces}



\icmlsetsymbol{equal}{*}

\begin{icmlauthorlist}
\icmlauthor{Victor Dorobantu}{caltech}
\icmlauthor{Charlotte Borcherds}{caltech}
\icmlauthor{Yisong Yue}{caltech}
\end{icmlauthorlist}

\icmlaffiliation{caltech}{Department of Computing and Mathematical Sciences, California Institute of Technology, Pasadena, CA, USA}

\icmlcorrespondingauthor{Victor Dorobantu}{vdoroban@caltech.edu}

\icmlkeywords{Machine Learning, Generative Modeling, Continuous Normalizing Flows, Computational Geometry, Conformal Geometry, Discrete Differential Geometry}

\vskip 0.3in
]



\printAffiliationsAndNotice{}  

\begin{abstract}
We propose \emph{conformal generative modeling}, a  framework for generative modeling on 2D surfaces approximated by discrete triangle meshes.  Our approach leverages advances in discrete conformal geometry to develop a map from a source triangle mesh to a target triangle mesh of a simple manifold such as a sphere.  After accounting for errors due to the mesh discretization, we can use any generative modeling approach developed for simple manifolds as a plug-and-play subroutine.  We demonstrate our framework on multiple complicated manifolds and multiple generative modeling subroutines, where we show that our approach can learn good estimates of distributions on meshes from samples, and can also learn simultaneously from multiple distinct meshes of the same underlying manifold.
\end{abstract}

\begin{figure*}
    \centering
    \includegraphics[width=\textwidth]{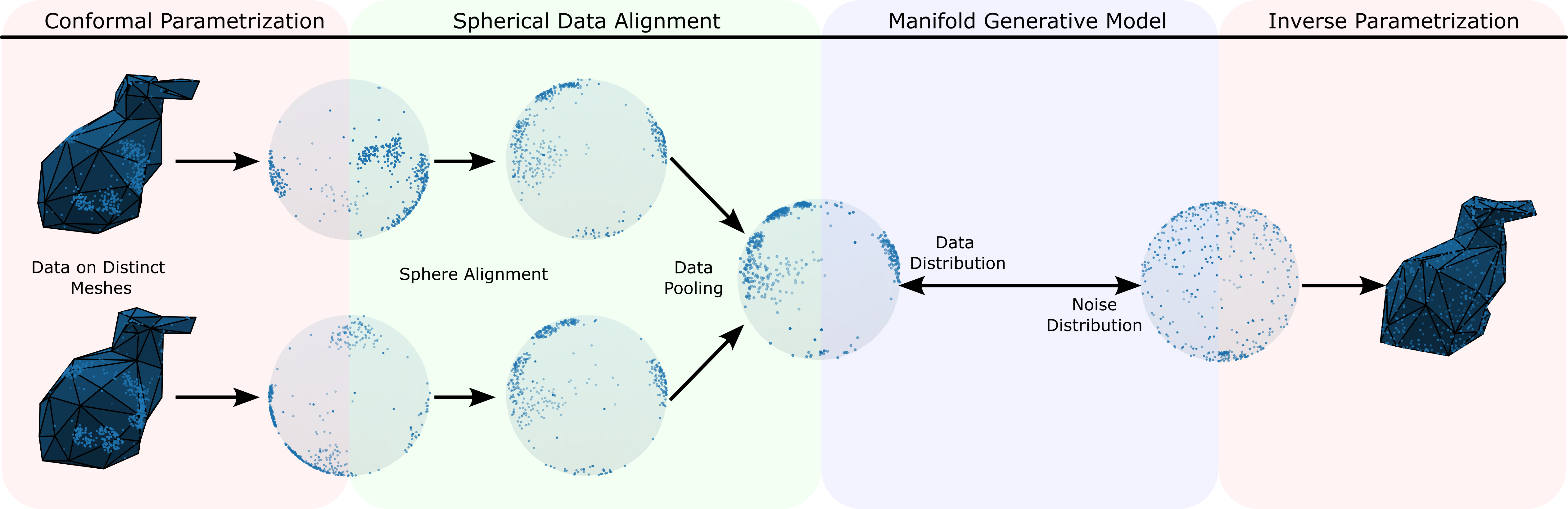}
    \vspace{-20pt}
    \caption{Overview of our approach. From left to right, original data samples are gathered from multiple meshes and mapped to unit sphere via discrete conformal transformations (red). The spheres are aligned with a reference mesh using rotations maximizing the correlation between log conformal factors, geometric signatures of the original meshes. Data is then aggregated on one sphere (green). A generative model transforms the data distribution on the sphere to a noise distribution on the sphere in the forward direction (blue). The resulting noise distribution is mapped to the reference mesh, again via a discrete conformal transformation. Here, the noise distribution on the reference mesh is uniform. Meshes are decimated to more clearly distinguish distinct meshes; for full meshes see \cref{fig:bunnies}.}
    \label{fig:overview}
\end{figure*}

\section{Introduction}
\label{sec:intro}

The study of learning expressive generative models has seen increasing interest in recent years, owing to the emergence of approaches such as continuous normalizing flows \citep{rezende2015variational,dinh2016density,papamakarios2021normalizing} and diffusion models \citep{sohl2015deep,song2020score}.   Such models have the ability to both efficiently draw samples from the learned distribution, as well as compute exact probabilities on the input space, making them useful for many applications in science and engineering from experiment design \citep{song2022general} to inverse problems \citep{gao2021deepgem,song2021solving}.

A key challenge in many domains is that the distribution to be estimated lies on a complicated manifold, rather than in Euclidean space.  Examples include modeling molecular activity \citep{chen2012triangulated,shapovalov2011smoothed}, robotic motion \citep{feiten2013rigid}, high energy physics \citep{brehmer2020flows},
and brain activity \citep{gerber2010manifold}.  Moreover, many domains use discrete approximations of the true manifold (e.g., triangle meshes). 

Our goal is to develop generative modeling approaches for complicated 2D surfaces in the case where the training data is represented using discrete mesh approximations.
Existing work on Riemannian generative modeling typically operate on canonical manifolds such as spheres, tori, hyperbolic spaces, and matrix Lie groups \citep{de2022riemannian,mathieu2020riemannian, lou2020neural, katsman2021equivariant} or else require uncontrolled approximations via learned implicit surface representations \citep{rozen2021moser}. The challenge of learning generative models directly on complicated meshes remains open.

\textbf{Our contributions.} We propose \emph{conformal generative modeling}, a framework for generative modeling on Riemannian manifolds approximated by discrete triangle meshes.  Our approach is based on establishing a conformal transformation between the source mesh and a target mesh that approximates a simple manifold such as a sphere \citep{springborn2008conformal}.  Such a transformation amounts to a diffeomorphism between the two manifolds, subject to accounting for approximation error from the mesh discretization.  Afterwards, one can use any  generative model for simple manifolds as a plug-and-play subroutine, which we demonstrate empirically on eight complicated manifolds (which is significantly more than has been demonstrated by other Riemannian generative modeling approaches). Our implementation is available at \href{https://github.com/vdorbs/spherical-generative-modeling.git}{github.com/vdorbs/spherical-generative-modeling.git}. 

A further benefit of conformal generative modeling is that it can learn simultaneously using data from multiple distinct meshes of the same underlying manifold.  This benefit comes from the ability to establish alignments between the conformal transformations of the meshes on the unit sphere (e.g., \citet{baden2018mobius,wang20063d}), as shown in Figure \ref{fig:overview}.  We demonstrate this ability empirically, usign data from multiple distinct meshes to train a single generative model capable of generating data on unseen meshes.



\section{Preliminaries \& Problem Statement}

Our approach uses topological equivalences to simple manifolds (e.g., closed 2D Riemannian manifolds with no ``holes'' are topological spheres). We first describe notation to describe such concepts and then the problem statement.

\subsection{Notation, Definitions, and Conventions}

\textbf{Basics.} We denote the unit two-sphere by $S^2$. For manifold $\mathcal{M}$, we denote its tangent space at any $p \in \mathcal{M}$ by $T_p\mathcal{M}$. For manifolds $\mathcal{M}$ and $\mathcal{N}$ and a smooth map $f: \mathcal{M} \to \mathcal{N}$, we denote the differential of $f$ at $p\in\mathcal{M}$ by $df_p: T_p\mathcal{M} \to T_{f(p)}\mathcal{N}$; this is a linear map between tangent spaces.

\textbf{Triangle Meshes \& Vertex Embeddings.}
A (Euclidean) triangle mesh is described by a triple $\mathsf{T} = (\mathsf{V}, \mathsf{E}, \mathsf{F})$ and a vertex embedding $\mathsf{f}: \mathsf{V} \to \R^d$, where $\mathsf{V}$ is the set of vertices, $\mathsf{E}$ is a set of ordered vertex pairs comprising edges, $\mathsf{F}$ is a set of ordered vertex triples comprising faces, and $\mathsf{f}_i \in \R^d$ denotes the position in space of vertex $i \in \mathsf{V}$. As a simplicial complex, for each face in $\mathsf{F}$, each pair of vertices must belong to the edge set $\mathsf{E}$, potentially with the vertex order reversed. In this paper, we will exclusively consider \textit{manifold} triangle meshes; in this case, each edge belongs to exactly two faces, with the exception of boundary edges (if they exist), which each belong to exactly one face. Such a triangle mesh is equipped with a \textit{discrete metric} (the discrete analogue of a Riemannian metric) $\ell: \mathsf{E} \to \R_{++}$:
\begin{equation}
    \ell_{ij} = \| \mathsf{f}_i - \mathsf{f}_j \|_2.
\end{equation}
This discrete metric must satisfy triangle inequality:
\begin{align}
    \ell_{ij} + \ell_{jk} &\leq \ell_{ki}, & \ell_{jk} + \ell_{ki} &\leq \ell_{ij}, & \ell_{ki} + \ell_{ij} &\leq \ell_{jk},
\end{align}
for each face $(i, j, k) \in \mathsf{F}$.

\textbf{Piecewise Linear Surfaces.} Often we will need to consider the piecewise linear surface generated from a vertex embedding by placing triangle vertices at the vertex positions of each face. We will typically denote such a surface as $\mathsf{M}$; we will assume available training data belongs to such surfaces.

\textbf{Topological Equivalences.}
The mesh $\mathsf{T}$ is a topological sphere if it has no boundary and has an Euler characteristic $| \mathsf{F} | - | \mathsf{E} | + | \mathsf{V} | = 2$ (i.e, has no ``holes''). This implies only that $\mathsf{T}$ has the connectivity of a sphere (topological information), though not necessarily the geometry of one. If $\mathsf{T}$ is a topological sphere \textit{and} $\mathsf{f}_i \in S^2$ for each vertex $i \in \mathsf{V}$, we will also consider \textit{spherical} triangle meshes, which partition the sphere into spherical triangles. 

A Riemannian manifold is a topological sphere if it is homeomorphic to a sphere; that is, if it admits a continuous bijection to the sphere with a continuous inverse.

\subsection{Problem Statement}

\begin{figure*}
    \centering
    \includegraphics[width=\textwidth]{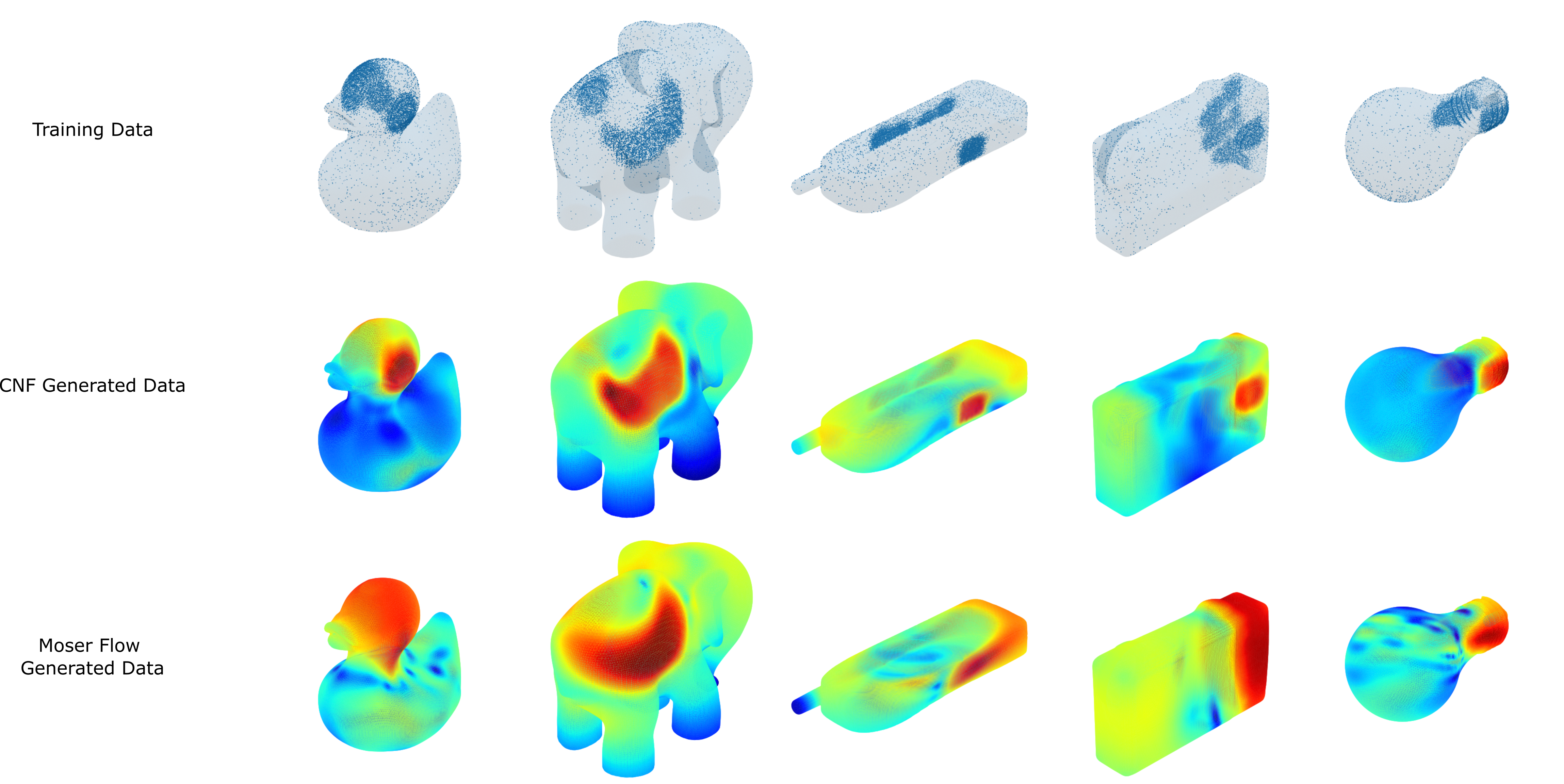}
    \vspace{-.15in}
    \caption{ Showing five meshes: training data (top row), learned CNF densities (middle row), and learned Moser flow densities (bottom row). From left to right, the meshes are \texttt{rubber-duck}, \texttt{elephant}, \texttt{cell-phone}, \texttt{camera}, and \texttt{light-bulb}.}
    \label{fig:densities}
\end{figure*}

Consider a 2D Riemannian manifold $\mathcal{M}$ that is a topological sphere. Our goal is to estimate a probability density function $\rho: \mathcal{M} \to \R_+$  using training samples generated i.i.d. from some underlying distribution on $\mathcal{M}$.  We are generally interested in complicated manifolds  (see Figures \ref{fig:overview} and \ref{fig:densities} for examples). Moreover, we assume that the training samples are represented on triangulated mesh approximations of $\mathcal{M}$.

Concretely, suppose we have access to $n$ triangle meshes $\mathsf{T}^{(1)}, \dots, \mathsf{T}^{(n)}$ that approximate $\mathcal{M}$, with respective vertex embeddings $\mathsf{f}^{(1)}, \dots, \mathsf{f}^{(n)}$ and piecewise linear surfaces $\mathsf{M}^{(1)}, \dots, \mathsf{M}^{(n)}$. We assume the vertex embeddings place the vertices on $\mathcal{M}$. We also have access to $n$ datasets, $\mathcal{D}^{(1)} \subset \mathsf{M}^{(1)}, \dots, \mathcal{D}^{(n)} \subset \mathsf{M}^{(n)}$, with each $\mathcal{D}^{(n)}$ comprising samples drawn i.i.d. from the underlying distribution on $\mathcal{M}$ but represented spatially on $\mathsf{M}^{(1)}, \dots, \mathsf{M}^{(n)}$. Figure \ref{fig:overview} (Left/Red) gives a depiction of the training data collected on two distinct meshes of the Stanford Bunny surface.


\section{Conformal Generative Modeling}

Our \emph{conformal generative modeling} framework is predicated on the idea of identifying an invertible transformation from each source mesh to the unit sphere.\footnote{One could in principle compute transformations to other simple manifolds with higher Euler numbers, such as tori.  However, certain algorithmic steps in the framework become more challenging, such as aligning the different meshes.} Given such a transformation, training samples from each mesh are effectively transformed to form a single pooled set of (weighted) training samples on the unit sphere.  Afterwards, one can use any existing Riemannian generative modeling approach estimate a probability density function on the sphere \citep{lou2020neural,mathieu2020riemannian,rozen2021moser,de2022riemannian}.One can then invert the transformation to produce outputs on any of the meshes.  Figure \ref{fig:overview} presents a high-level depiction of our framework.

We first summarize the key steps, and then develop the necessary technical tools on conformal geometry, spherical parameterizations, and M\"obius registration (Sections \ref{sec:conformal} \& \ref{sec:area}) to describe the full technical  details of the approach (Section \ref{sec:solution} and Algorithm \ref{algo:main}).

\textbf{Conformal Transformations \& Conformal Factors.}
The specific type of transformations we will use are \emph{discrete conformal transformations} \citep{springborn2008conformal}. Informally, conformal transformations are maximally angle-preserving\footnote{Complete angle preservation overly restricts the class of admissible transformations, yielding only rigid isometries. See \citet{Crane:2020:DCG} for further discussion.} maps between two meshes (e.g., the source manifold and the sphere).  In other words, for any two edges that share a vertex in common, the angle between those two edges are preserved as much as possible.  A direct consequence is that the shape of each face on the mesh is also approximately preserved in the transformation. Importantly, however, conformal transformations may distort differential area. In doing so, a conformal transformation rescales edge lengths via a \emph{conformal factor} at each vertex.

 \textbf{Aligning Multiple Meshes.} In order to pool training data from multiple meshes, it is necessary to align their transformed meshes on the sphere, which amounts to finding a rotation of each mesh (Section \ref{sec:registration}). This step is skipped in the special case where we learn with only one mesh. 

\textbf{Creating Training Samples on the Sphere.}  In order to generate proper training examples on the sphere, two additional steps are needed.  First, the conformal transformation has stretched or shrunk various parts of the source mesh, and so one must re-weight training samples in order to transport the original measure to the sphere.  Intuitively, data points in stretched portions should receive higher weight, and those in shrunk portions should receive lower weight.

\begin{figure}
    \centering
    \includegraphics[width=0.5\textwidth]{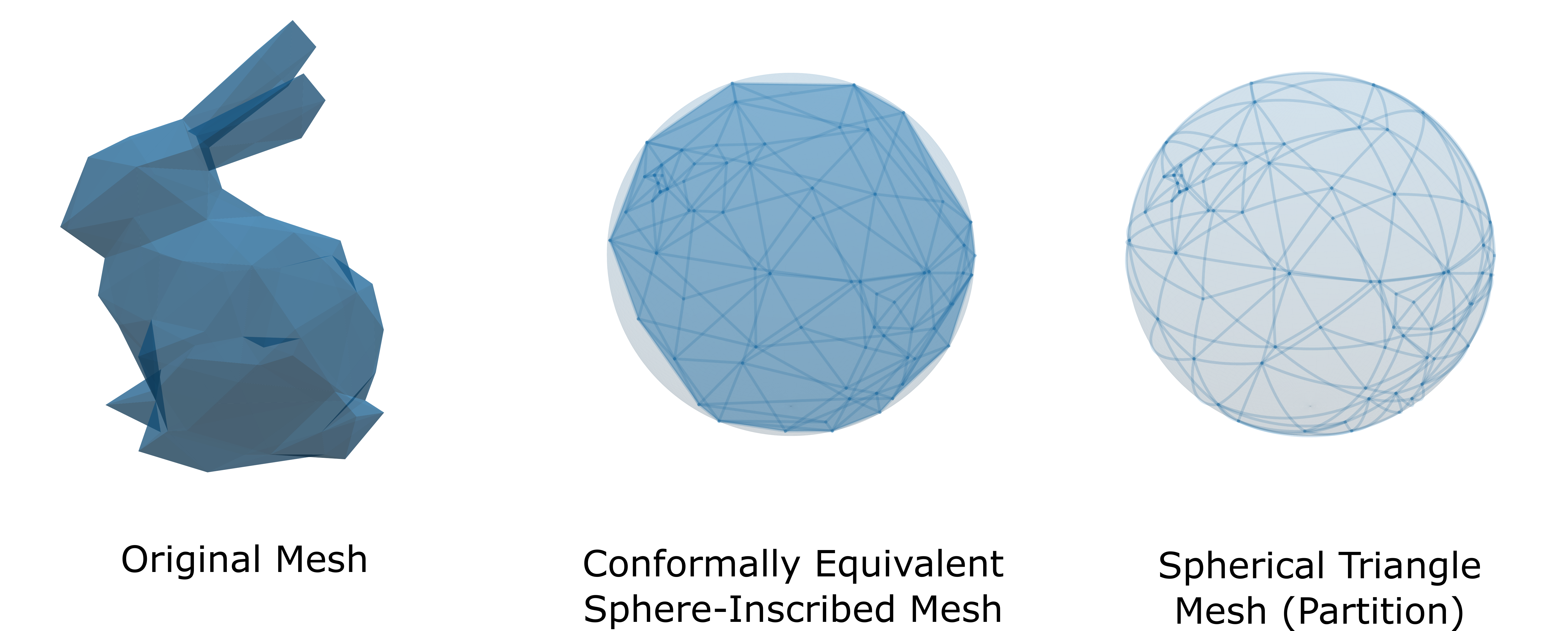}
    \vspace{-20pt}
    \caption{Correspondence between piecewise linear surfaces and unit sphere, with the original mesh on the left, the mesh after spherical parametrization in the center (shown inscribed within the unit sphere), and the associated partition of the sphere into spherical triangles on the right.}
    \label{fig:correspondence}
\end{figure}

The second step is to transform the linear triangle surfaces into spherical triangle surfaces.  Figure \ref{fig:correspondence} depicts this step, where the middle is a piecewise linear mesh that is circumscribed by the unit sphere (all the edges lie inside the sphere), and the right is a conversion of each triangle surface to a spherical one.  It is straightforward to compose this transformation with the conformal transformation. In Section \ref{sec:spherical}, we describe spherical parameterization, and in Section \ref{sec:area} we derive the additional differentiable change in area to compute an additional weighting factor for the training samples when transforming to a spherical mesh.

\begin{algorithm}[t]
	\begin{small}
	\caption{Conformal Generative Modeling} 
	\label{algo:main}
	\begin{algorithmic}[1]
            \INPUT Triangle meshes $\mathsf{T}^{(1)}, \dots, \mathsf{T}^{(n)}$ and correpsonding piecewise linear surfaces $\mathsf{M}^{(1)}, \dots, \mathsf{M}^{(n)}$
            \INPUT Training sets $\mathcal{D}^{(1)} \subset \mathsf{M}^{(1)}, \dots, \mathcal{D}^{(n)} \subset \mathsf{M}^{(n)}$
            \STATE Compute conformal transformation from each mesh to the unit sphere (\cref{sec:spherical}), obtaining $\mathsf{M}_{\mathrm{inscr}}^{(1)}, \dots, \mathsf{M}_{\mathrm{inscr}}^{(n)}$
            \STATE Align $\mathsf{M}^{(2)}_{\mathrm{inscr}}, \dots, \mathsf{M}^{(n)}_{\mathrm{inscr}}$ to $\mathsf{M}^{(1)}_{\mathrm{inscr}}$ with maximum correlation rotations (\cref{eqn:correlation})
            \STATE Map data to aligned spheres, obtaining $\tilde{\mathcal{D}}^{(1)}, \dots, \tilde{\mathcal{D}}^{(n)}$ 
            \STATE Compute log changes of area from projection (\cref{eqn:change-of-area}) and triangle area ratios for each data point in each of $\tilde{\mathcal{D}}^{(1)}, \dots, \tilde{\mathcal{D}}^{(n)}$
            \STATE Train likelihood-based generative model on sphere, subtracting log change of area from log probability densities in loss function (\cref{eqn:data-likelihood})
            \STATE Return sphere generative model and inverse conformal transformations mapping to each of $\mathsf{M}^{(1)}, \dots, \mathsf{M}^{(n)}$
 	\end{algorithmic}
  \end{small}
\end{algorithm}

\subsection{Background on Conformal Geometry}
\label{sec:conformal}

\subsubsection{Conformal Equivalence}

Conformal maps locally preserve angles and orientations. Let $\mathcal{M}$ and $\tilde{\mathcal{M}}$ be Riemannian manifolds with corresponding Riemannian metrics $g$ and $\tilde{g}$, respectively. A smooth local embedding $f: \mathcal{M} \to \tilde{\mathcal{M}}$ is \textit{conformal} if it preserves orientations and there exists a  $u: \mathcal{M} \to \R$ satisfying:
\begin{equation}
    \tilde{g}_{f(p)}(df_p(v), df_p(w)) = e^{2u(p)} \cdot g_p(v, w),
\end{equation}
for all points $p \in \mathcal{M}$ and tangent vectors $v, w \in T_p\mathcal{M}$. The function $u$ is called the \textit{log conformal factor}. Intuitively, we can equivalently push forward $v$ and $w$ via the differential $df_p$ and compute their inner product in the tangent space at $f(p)$, or we can compute the inner product of $v$ and $w$ in the tangent space at $p$ and scale the product by $e^{2u(p)}$.

Analogously, consider a Euclidean triangle mesh $\textsf{T} = (\mathsf{V}, \mathsf{E}, \mathsf{F})$ with two embeddings $\mathsf{f}$ and $\tilde{\mathsf{f}}$ (e.g., one for the source manifold and one for the sphere), corresponding to discrete metrics $\ell$ and $\tilde{\ell}$, respectively. The metrics are \textit{(discretely) conformally equivalent} \cite{springborn2008conformal} if there is a function $u: \mathsf{V} \to \R$ satisfying:
\begin{equation}
    \tilde{\ell}_{ij} = e^{(u_i + u_j) / 2} \cdot \ell_{ij},
\end{equation}
for all edges $(i, j) \in \mathsf{E}$. As in the continuous case, the function $u$ is also called the log conformal factor. Intuitively, $e^{2u_i} \in \R_{++}$ captures how differential area changes (multiplicatively) as the vertex $i \in \mathsf{V}$ is sent from $\mathsf{f}_i$ to $\tilde{\mathsf{f}}_i$.

Two important properties of discrete conformal equivalence are transitivity and invariance under M\"obius transformations. First, suppose $\ell_1$, $\ell_2$, and $\ell_3$ are discrete metrics on the same triangle mesh. If $\ell_1$ and $\ell_2$ are discretely conformally equivalent with log conformal factor $u_1$ and $\ell_2$ and $\ell_3$ are discretely conformally equivalent with log conformal factor $u_2$, then $\ell_1$ and $\ell_3$ are discretely conformally equivalent with log conformal factor $u_1 + u_2$. Second, if two vertex embeddings corresponding to the same mesh are related via a M\"obius transformation (which includes Euclidean transformations, sphere inversions, and stereographic projections), then the corresponding discrete metrics are discretely conformally equivalent \cite{springborn2008conformal}.

\subsubsection{Spherical Parametrizations}
\label{sec:spherical}

\begin{figure}
    \centering
    \includegraphics[width=0.25\textwidth]{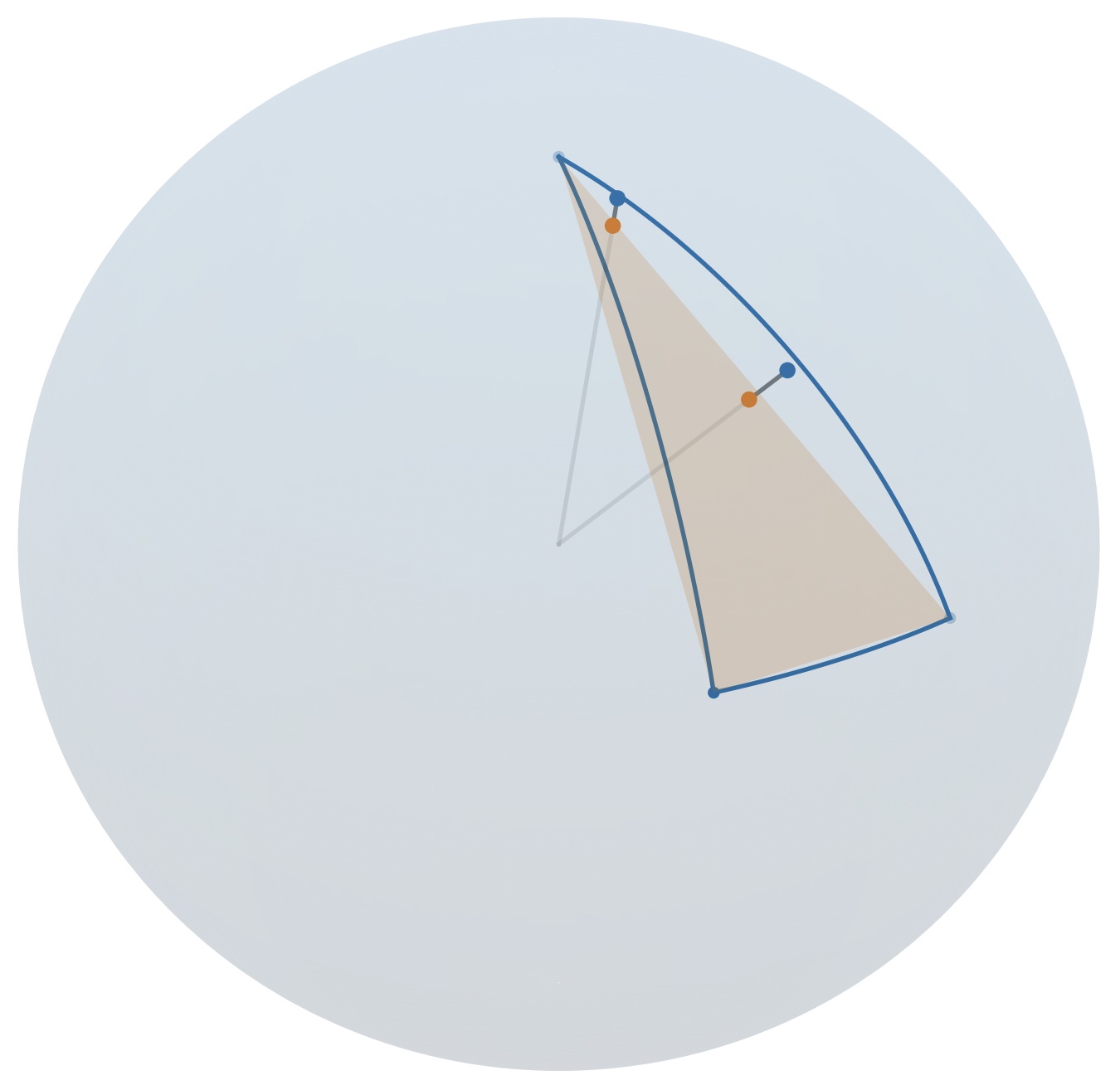}
    \includegraphics[width=0.2\textwidth]{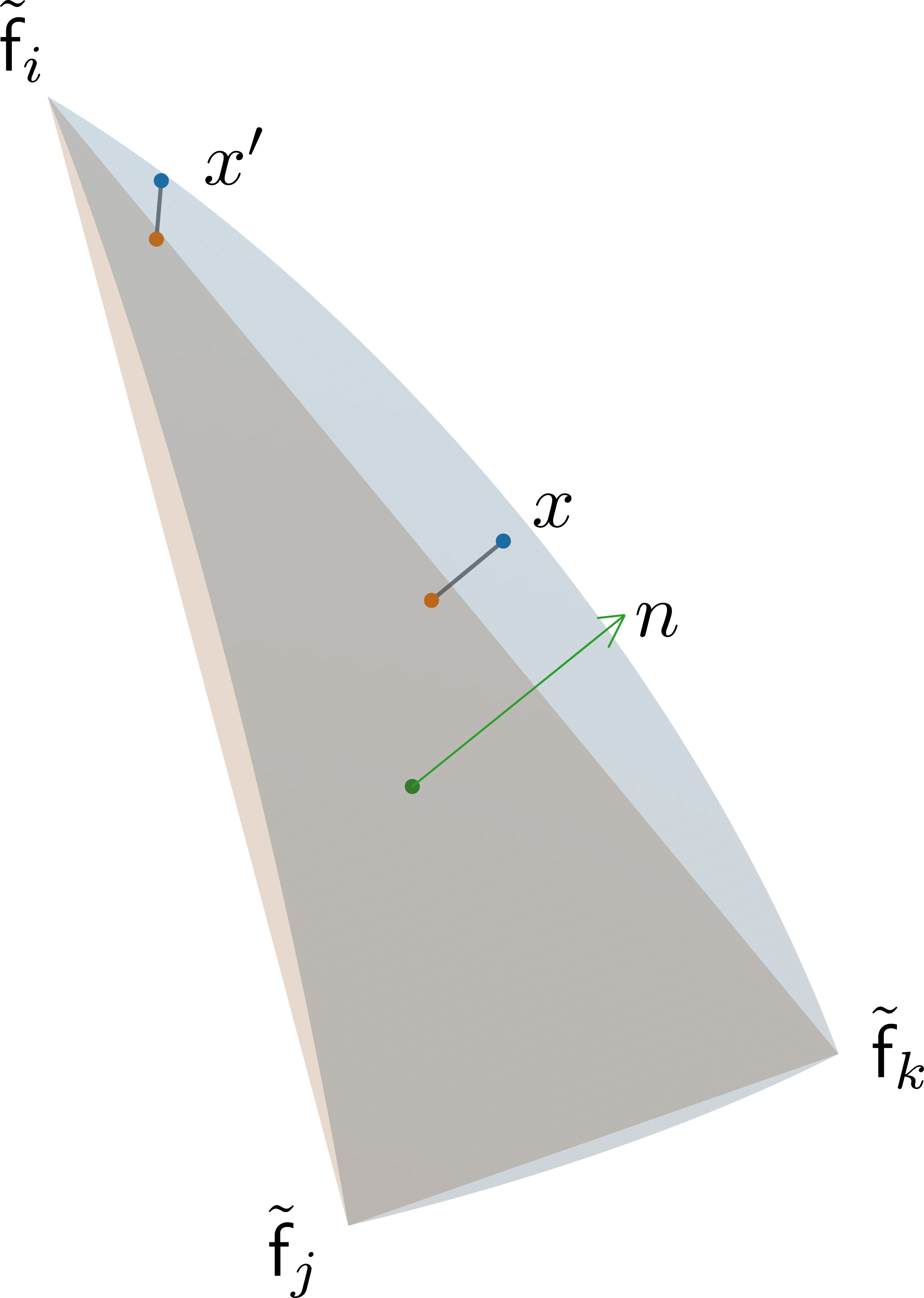}
    \caption{Corrsepondence between an inscribed Euclidean triangle and its associated spherical triangle. Points on the sphere are  in blue, projected points (onto the inscribed triangles) are  in orange, and the inscribed triangle normal vector is  in green.}
    \label{fig:projection}
\end{figure}

Consider a triangle mesh $\mathsf{T} = (\mathsf{V}, \mathsf{E}, \mathsf{F})$ that approximates a sphere, with vertex embedding $\mathsf{f}$ and discrete metric $\ell$. Typically, such a mesh is circumscribed by the sphere, meaning the vertices lie on the sphere and the edges lie within.  Our goal here is to transform each linear triangle surface in $\mathsf{T}$ into a spherical triangle (Figure \ref{fig:projection}). 
We use the following procedure \cite{springborn2008conformal,bobenko2016discrete}, which returns a new embedding $\tilde{\mathsf{f}}$ of $\mathsf{T}$ that maps the vertices to the sphere $S^2$ and corresponds to a discrete metric that is discretely conformally equivalent to $\ell$:
\begin{enumerate}
    \item Select an arbitrary vertex $i^* \in \mathsf{V}$ to be removed,
    \vspace{-2pt}
    \item Apply a change of discrete metric, making each neighbor of $i^*$ equally distant from $i^*$ (this metric is discretely conformally equivalent to $\ell$),
    \vspace{-2pt}
    \item Remove $i^*$ from $\mathsf{V}$, and all incident edges from $\mathsf{E}$ and  incident faces from $\mathsf{F}$ (yielding a topological disk),
    \vspace{-2pt}
    \item Compute a \textit{flat} discretely conformally equivalent metric via convex optimization which leaves the distances between boundary vertices unchanged,
    \vspace{-2pt}
    \item Embed the vertices in the plane $\R^2$ with edge lengths determined by the new flat metric,
    \vspace{-2pt}
    \item Stereographically project the vertices onto the unit sphere $S^2$ (through the north pole), maintaining discrete conformal equivalence of metrics,
    \vspace{-2pt}
    \item Reinsert $i^*$ at the north pole, along with removed edges and faces,
    \vspace{-2pt}
    \item Apply M\"obius transformations that map the sphere to itself (Lorentz transformations) to move the center of the vertex positions to the origin, again maintaining discrete equivalence of metrics.
\end{enumerate}

Through spherical parametrization, we establish correspondences between three surfaces: a piecewise linear surface described by $\mathsf{f}$, a piecewise linear surface inscribed within the unit sphere and described by $\tilde{\mathsf{f}}$, and the unit sphere itself. These correspondences are illustrated in \cref{fig:correspondence}. Appendix \ref{sec:spherical2} provides the full details of this procedure.  The key ideas are to: (i) use a convex energy formulation to compute the flat metric  (Step 4), and (ii) produce an interpolation of the conformal factor on the sphere.

\subsubsection{M\"obius Registration}
\label{sec:registration}

In order to completely relate two meshes of the same manifold (e.g., to pool their training data), we must align their  representations.
With the  result of Section \ref{sec:spherical}, two spherical parametrizations of the same mesh will differ only by a rotation \cite{baden2018mobius}. The set of rotations is substantially easier to work with than the set of Lorentz transformations. This rotation can be found via harmonic analysis.

For two complex-valued functions on the sphere $f, g: S^2 \to \C$, we can define a correlation function $C: SO(3) \to \C$ on the rotation group $SO(3)$ as:
\begin{equation}\label{eqn:correlation}
    C(R) = \int_{S^2} f(x)\overline{g(R^\top x)}~\mathrm{d}x,
\end{equation}
for all rotation matrices $R \in SO(3)$. If $f$ and $g$ are real-valued, then so is $C$, and we can seek its global maximizer. The spherical harmonic transform \cite{kostelec2008ffts} yields coefficients such that:
\begin{align}
    f(x) &= \sum_{l = 0}^\infty \sum_{m = -l}^l \hat{f}_{lm} Y_l^m(x),\nonumber\\
    g(x) &= \sum_{l = 0}^\infty \sum_{m = -l}^\ell \hat{g}_{lm} Y_l^m(x),
\end{align}
for all $x \in S^2$, where $Y_l^m: S^2 \to \C$ denotes the spherical harmonic of $l$ and order $m$. The $SO(3)$ Fourier transform of $C$ can then be written in terms of the spherical harmonic transforms of $f$ and $g$, as:
\begin{equation}\label{eqn:SO3-fft}
    C(R) = \sum_{l = 0}^\infty \sum_{m = -l}^l \sum_{m' = -l}^l \hat{f}_{lm} \overline{\hat{g}_{lm'}} \cdot \overline{D_{mm'}^l(R)},
\end{equation}
for all rotations $R \in SO(3)$, where $D_{mm'}^l$ denotes a Wigner $D$-function \cite{kostelec2008ffts}. Intuitively, the coefficients of \cref{eqn:SO3-fft} are degree-wise outer products of the coefficients comprising each degree of the spherical harmonic transforms of $f$ and $g$.

While evaluating $C$ at any rotation is computationally difficult, \citet{kostelec2008ffts} provide a fast method using a grid of Euler angles. This method makes use of the fast spherical harmonic transform, sampling $f$ and $g$ onto a spherical grid. We found a $64 \times 64$ spherical grid (a bandwidth of $32$) to be sufficient for our experiments.

\subsection{Differential Change of Area}
\label{sec:area}

For $l \in \{ 1, \dots, n \}$, consider a data point $x \in \tilde{\mathcal{D}}^{(l)}$. The data point $x$ is mapped from the sphere to $\mathsf{M}^{(l)}$ (the inverse transformation) in three stages; first $x$ is mapped onto a Euclidean triangle of $\tilde{\mathsf{M}}_{\mathrm{inscr}}^{(l)}$, then rotated by the inverse rotation ${R^{(l)}}^\top$ to lie on $\mathsf{M}_{\mathrm{inscr}}^{(l)}$, and finally moved to the point on $\mathsf{M}^{(l)}$ with the same barycentric coordinates in the corresponding face. The rotation is isometric, incurring no change of area, and the final map from $\mathsf{M}_{\mathrm{inscr}}^{(l)}$ to $\mathsf{M}^{(l)}$ is piecewise linear, incurring a multiplicative change of area equal to the area ratio of corresponding triangles (original triangle area divided by inscribed triangle area).

The first mapping, however, is nonlinear. To compute the corresponding change of area, suppose $x$ is in the spherical triangle characterized by $(i, j, k) \in \mathsf{F}^{(l)}$, where $\mathsf{F}^{(l)}$ is the set of faces in $\mathsf{T}^{(l)}$. Let $n$ denote the normalized cross product of edge vectors in the face (yielding a unit normal vector), with:
\begin{equation}
    n = \frac{(\tilde{\mathsf{f}}_j^{(l)} - \tilde{\mathsf{f}}_i^{(l)}) \times (\tilde{\mathsf{f}}_k^{(l)} - \tilde{\mathsf{f}}_i^{(l)})}{\| (\tilde{\mathsf{f}}_j^{(l)} - \tilde{\mathsf{f}}_i^{(l)}) \times (\tilde{\mathsf{f}}_k^{(l)} - \tilde{\mathsf{f}}_i^{(l)}) \|_2}.
\end{equation}
From \cref{eqn:scale-factor}, we find that $x$ is mapped to:
\begin{equation}
    y = \frac{n^\top \tilde{\mathsf{f}}_i^{(l)}}{n^\top x}\cdot x,
\end{equation}
with Jacobian matrix given as:
\begin{equation}
    \frac{\mathrm{d}y}{\mathrm{d}x} = \frac{(n^\top x)(n^\top \tilde{\mathsf{f}}_i^{(l)})I_3 - (n^\top \tilde{\mathsf{f}}_i^{(l)})xn^\top}{(n^\top x)^2}.
\end{equation}
Now, consider any tangent vector to the sphere $v \in \R^3$, with $x^\top v = 0$. Suppose additionally that $v$ has unit magnitude. The cross product $x \times v$ is another tangent vector, the result of rotating $v$ in the tangent plane counterclockwise by a quarter turn. The pair $(v, x \times v)$ form an orthonormal basis for the tangent plane, and the triple $(v, x \times v, x)$ forms a right-handed orthonormal basis for $\R^3$. The parallelogram spanned by $v$ and $x \times v$ is a square with area $1$, so we can compute the multiplicative change of area incurred when mapping $x$ to $y$ simply by computing the area of the parallelogram obtained by pushing forward $v$ and $x \times v$ by the Jacobian $\mathrm{d}y / \mathrm{d}x$. This area is:
\begin{align}\label{eqn:change-of-area}
    & \left\| \left(\frac{\mathrm{d}y}{\mathrm{d}x} \cdot v\right) \times \left(\frac{\mathrm{d}y}{\mathrm{d}x} \cdot (x \times v) \right) \right\|_2\nonumber\\
    &~= \frac{(n^\top \tilde{\mathsf{f}}_i^{(l)})^2}{(n^\top x)^4} \cdot \big\| \left((n^\top x)v - (n^\top v)x\right) \nonumber\\
    &~\qquad\qquad\qquad \times \left( (n^\top x)(x \times v) - n^\top (x \times v)x \right) \big\|_2\nonumber\\
    &~= \frac{(n^\top \tilde{\mathsf{f}}_i^{(l)})^2}{|n^\top x|^3} \cdot \big\| (x^\top n)x + ((x \times v)^\top n)(x \times v) \nonumber\\
    &~\qquad\qquad\qquad+ (v^\top n)v \big\|_2\nonumber\\
    &~= \frac{(n^\top \tilde{\mathsf{f}}_i^{(l)})^2}{|n^\top x|^3}.
\end{align}
The last equality follows since the term inside the norm is just the sum of the projections of $n$ onto $x$, $x \times v$, and $v$; since these three vectors form an orthonormal basis, the resulting sum has unit norm. Since:
\begin{equation}
    ((\tilde{\mathsf{f}}_j^{(l)} - \tilde{\mathsf{f}}_i^{(l)}) \times (\tilde{\mathsf{f}}_k^{(l)} - \tilde{\mathsf{f}}_i^{(l)}))^\top \tilde{\mathsf{f}}_i^{(l)} = (\tilde{\mathsf{f}}_j^{(l)} \times \tilde{\mathsf{f}}_k^{(l)})^\top \tilde{\mathsf{f}}_i^{(l)},
\end{equation}
we can write the final expression in \cref{eqn:change-of-area} as:
\begin{equation}\label{eqn:change-of-area-first-rewrite}
    \frac{((\tilde{\mathsf{f}}_j^{(l)} \times \tilde{\mathsf{f}}_k^{(l)})^\top \tilde{\mathsf{f}}_i^{(l)})^2 \cdot \| (\tilde{\mathsf{f}}_j^{(l)} - \tilde{\mathsf{f}}_i^{(l)}) \times (\tilde{\mathsf{f}}_k^{(l)} - \tilde{\mathsf{f}}_i^{(l)}) \|_2}{|((\tilde{\mathsf{f}}_j^{(l)} - \tilde{\mathsf{f}}_i^{(l)}) \times (\tilde{\mathsf{f}}_k^{(l)} - \tilde{\mathsf{f}}_i^{(l)}))^\top x|^3},
\end{equation}
where the norm in the numerator is the area of the inscribed triangle characterized by $(i, j, k)$.

To summarize, the differential change of area when mapping $x$ from the sphere to $\mathsf{M}^{(l)}$ is, as a multiplicative factor, the product of \cref{eqn:change-of-area-rewrite} and the ratio of areas for face $(i, j, k)$, represented both in $\mathsf{M}^{(l)}$ and $\mathsf{M}_{\mathrm{inscr}}^{(l)}$. Using the observation about the norm from \cref{eqn:change-of-area-first-rewrite}, this differential change of area is:
\begin{equation}\label{eqn:change-of-area-rewrite}
    \Delta^{(\ell)} \triangleq \frac{((\tilde{\mathsf{f}}_j^{(l)} \times \tilde{\mathsf{f}}_k^{(l)})^\top \tilde{\mathsf{f}}_i^{(l)})^2 \| (\mathsf{f}_j^{(l)} - \mathsf{f}_i^{(l)}) \times (\mathsf{f}_k^{(l)} - \mathsf{f}_i^{(l)}) \|_2}{|((\tilde{\mathsf{f}}_j^{(l)} - \tilde{\mathsf{f}}_i^{(l)}) \times (\tilde{\mathsf{f}}_k^{(l)} - \tilde{\mathsf{f}}_i^{(l)}))^\top x|^3},
\end{equation}
where now the norm in the numerator is instead the area of the triangle characterized by $(i, j, k)$ in the original mesh.

\subsection{Technical Description of Our Solution}
\label{sec:solution}

We can now fully instantiate Algorithm \ref{algo:main}.  In Step 1,  we spherically parametrize each of the piecewise linear surfaces, obtaining piecewise linear surfaces $\mathsf{M}_{\mathrm{inscr}}^{(1)}, \dots, \mathsf{M}_{\mathrm{inscr}}^{(n)}$ inscribed within the unit sphere and log conformal factor interpolations $\bar{u}^{(1)}, \dots, \bar{u}^{(n)}: S^2 \to \R$. For $i \in \{ 2, \dots, n \}$, we approximate the maximum correlation rotation matrix $R^{(i)} \in SO(3)$, maximizing the correlation function:
\begin{equation}
    C^{(i)}(R) = \int_{S^2} \bar{u}^{(1)}(x) \bar{u}^{(i)}(R^\top x) ~\mathrm{d}x,
\end{equation}
over all rotation matrices $R \in SO(3)$. With identity rotation $R^{(1)} = I_3$, we rotate each of the inscribed surfaces by the corresponding rotation, obtaining rotated surfaces $\tilde{\mathsf{M}}_{\mathrm{inscr}}^{(1)}, \dots, \tilde{\mathsf{M}}_{\mathrm{inscr}}^{(n)}$. In Step 2,  we map each data point in each $\mathcal{D}^{(i)}$ to $\tilde{\mathsf{M}}_{\mathrm{inscr}}^{(i)}$ and normalize the result to lie on the unit sphere, obtaining a spherical dataset $\tilde{\mathcal{D}}^{(i)}$.

We can now aggregate the spherical datasets and train a generative model using the resulting dataset (Step 3). However, to train such models with maximum likelihood estimation on the original surfaces $\mathsf{M}^{(1)}, \dots, \mathsf{M}^{(n)}$, we require the corresponding changes of area incurred by each of the maps to the unit sphere (Step 4). When evaluating log probability densities via a generative model on the sphere, we must subsequently subtract the log changes of area to compute log probability densities (corresponding to dividing density by change of area) on the original surface.

Elaborating on this point, suppose we choose a generative model on the sphere with density $\rho: S^2 \to \R_+$. For a dataset $\tilde{\mathcal{D}}^{(1)} \cup \dots \cup \tilde{\mathcal{D}}^{(n)}$ comprised of aligned spherical datasets, the data log likelihood (under the spherical density) is:
\begin{equation}
    \sum_{l = 1}^n \sum_{x \in \tilde{\mathcal{D}}^{(l)}} \log \rho(x).
\end{equation}
while the corrected data log likelihood (under the corresponding mesh densities) is simply:
\begin{equation}\label{eqn:data-likelihood}
    \sum_{l = 1}^n \sum_{x \in \tilde{\mathcal{D}}^{(l)}} \left( \log{\rho(x)} - \log{\Delta^{(l)}(x)} \right),
\end{equation}
where $\Delta^\ell(x)$ denotes the change of area from \cref{eqn:change-of-area-rewrite} for data point $x \in \tilde{\mathcal{D}}^{(l)}$. Note that the change of area terms only need to be computed once at the start of training for such a generative model via likelihood maximization; the only terms which cannot be precomputed include $\rho$, as this density changes throughout training.

The remaining Step 5 is to choose an appropriate generative model to train on the sphere, which we discuss next.

\subsection{Choosing a Base Generative Model}
\label{sec:base}

 Step 5 in Algorithm \ref{algo:main}, in principle one could choose any Riemannian generative model that trains on canonical manifolds such as the unit sphere.  In this paper, we instantiate our framework using two approaches: Riemannian continuous normalizing flows (CNFs) \cite{mathieu2020riemannian} and Moser flows \cite{rozen2021moser}, which we discuss in detail in Section \ref{sec:base2}.

 One issue we encountered is determining a good projection of a Euclidean gradient (which is the native gradient representation of many flow-based generative models) onto the sphere.  A bad projection can lead to numerical instability and longer computational costs.  For all of our base generative models, we adopt the approach of \citet{rozen2021moser} using a normalizing trick discussed in Section \ref{sec:CNF}.

\section{Experiments}


We first demonstrate our method without the need for spherical alignment. We demonstrate our approach on 8 different meshes (see \cref{tab:results}), which is significantly more than has been empirically demonstrated by other Riemannian generative modeling approaches. We train continuous normalizing flows and Moser flows on each mesh. The meshes and data used are from the ContactDB dataset \cite{brahmbhatt2019contactdb}. Second, we show how our approach enables the sharing of data from distinct meshes approximating the same underlying manifold. We generate $5$ similar but distinct meshes (derived from \texttt{stanford-bunny}) to show how a single generative model can be trained using several meshes corresponding to the same underlying manifold. We show that log likelihood tested on a held-out mesh improves with additional data, even when the data comes from different meshes. We also show how all $5$ meshes can be used to simultaneously generate data on each of the $5$ meshes. All experiments were run on a single NVIDIA RTX A6000 GPU. In all experiments, noise distributions are uniform on the sphere. Our code is available at \href{https://github.com/vdorbs/spherical-generative-modeling.git}{github.com/vdorbs/spherical-generative-modeling.git}

\subsection{Training Data}\label{sec:contactdb}

ContactDB \cite{brahmbhatt2019contactdb} is a collection of grasping and manipulation data for applications in human-robot interaction. The data were collected by thermally imaging 3D printed household objects after they were grasped by human subjects. We study $8$ objects, each handled by a single participant. The data take the form of triangle meshes with a single feature (intensity) at each vertex, indicating thermal energy transferred to the object. As in \cite{brahmbhatt2019contactdb}, we process the contact maps by passing the intensities through a sigmoid function, assigning $0.999$ probability to the maximum intensity and $0.001$ to the minimum. As in \cite{rozen2021moser}, we average the corresponding probabilities on each face to obtain an unnormalized probability distribution over faces. We then draw $10000$ samples from each mesh, each sample generated by sampling a face according to the unnormalized distribution, then by sampling a uniformly distributed point on the sampled face.

\subsection{Distribution Modeling Results}\label{sec:dist-modeling}

We run $5$ trials for each of the $8$ meshes, training both a CNF and a Moser flow. Validation results are listed in \cref{tab:results}. In both cases we use $5000$ training samples and $5000$ validation samples. The CNF vector field and Moser flow flux field are both parametrized as $3$-hidden layer neural networks with hidden dimensions of $32$ and $\tanh$ nonlinearities. The inputs to both neural networks are vectors in $\R^3$, with the CNF taking a fourth input of time. Both models are trained with the Adam optimizer \cite{kingma2014adam}. The CNF is trained for $100$ epochs with a batch size of $256$ and a learning rate of $10^{-2}$. The Moser flow is trained for $4000$ epochs with a batch size of $256$ and a learning rate of $10^{-4}$. These results show that our approach can reliably model distributions on multiple manifolds using multiple Riemannian generative models as subroutines.

\begin{table}[t]
\caption{Log likelihoods for $8$ meshes (higher is better), averaged over $5$ runs per mesh, plus or minus one standard deviation.\label{tab:results}}
\label{sample-table}
\vspace{-0.1in}
\begin{center}
\begin{small}
\begin{sc}
\begin{tabular}{lcc}
\toprule
Mesh & CNF & Moser Flow \\
\midrule
\texttt{camera}         & $4 \pm 0.01703$      & $3.726 \pm 0.1048$ \\
\texttt{stanford-bunny} & $4.354 \pm 0.0435$   & $4.168 \pm 0.1937$ \\
\texttt{light-bulb}     & $4.44 \pm 0.04114$   & $4.239 \pm 0.07521$ \\
\texttt{elephant}       & $4.472 \pm 0.01215$  & $4.462 \pm 0.0361$ \\
\texttt{mouse}          & $4.769 \pm 0.009795$ & $4.726 \pm 0.06756$ \\
\texttt{cell-phone}     & $5.465 \pm 0.01964$  & $5.302 \pm 0.1207$ \\
\texttt{rubber-duck}    & $5.75 \pm 0.007362$  & $5.368 \pm 0.1799$ \\
\texttt{banana}         & $6.591 \pm 0.02889$  & $6.318 \pm 0.1723$ \\
\bottomrule
\end{tabular}
\end{sc}
\end{small}
\end{center}
\vskip -0.1in
\end{table}

\subsection{Spherical Alignment}

\begin{figure*}
    \centering
    \includegraphics[width=0.9\textwidth]{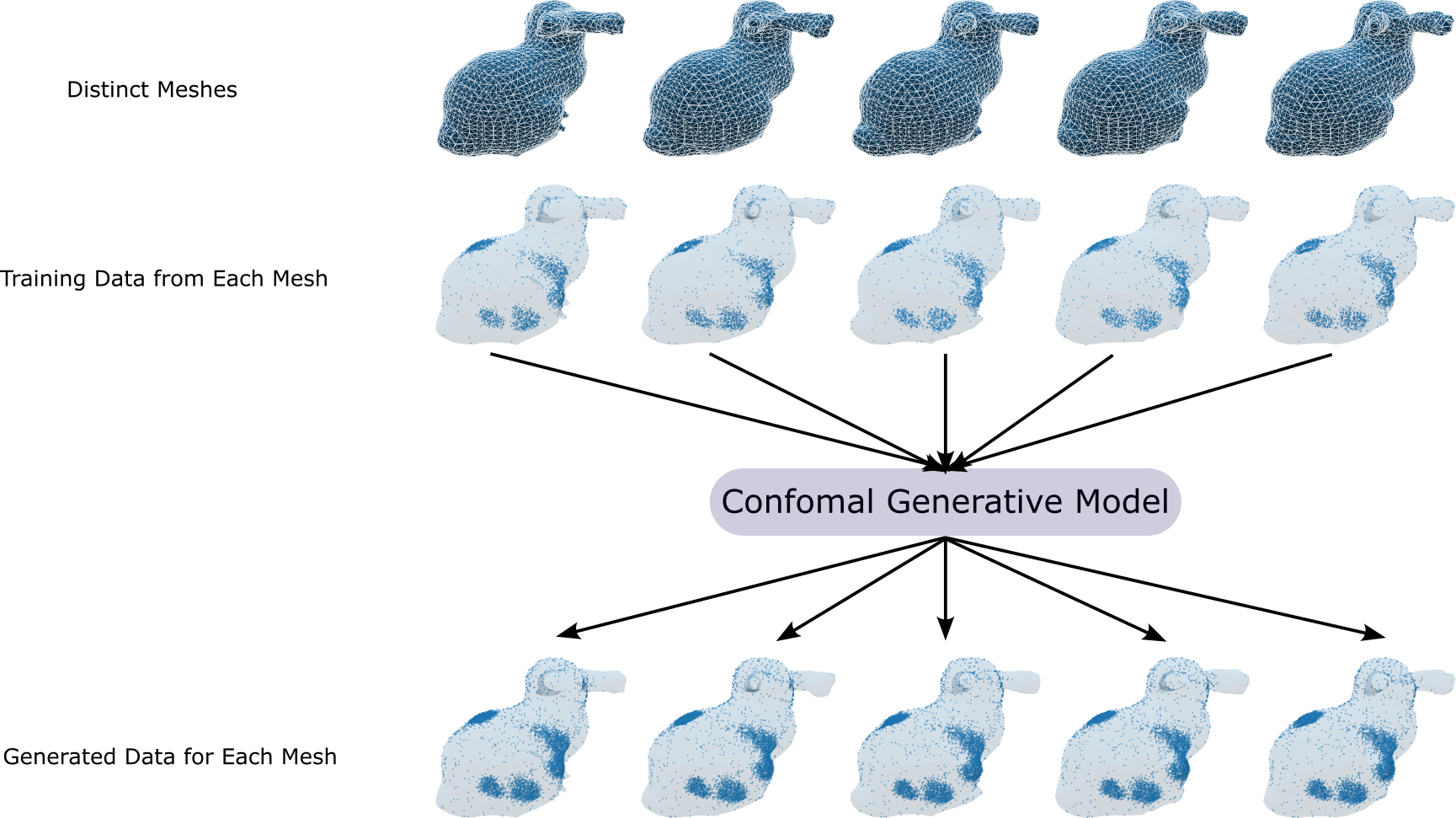}
    \caption{Distinct meshes (top row), training data on each mesh (second row) and samples generated from learned model (bottom row). The same model is used to generate data on each of the five meshes, trained using data aggregated from all meshes.}
    \label{fig:bunnies}
\end{figure*}

We next demonstrate our method when we have access to data on distinct meshes. For a single mesh, we sample data as in \cref{sec:contactdb}. We generate $5$ distinct meshes by sampling $5$ sets of vertices at random, each with only $1\%$ of the vertices in the original mesh. The meshes are then obtained from Poisson surface reconstruction \cite{kazhdan2006poisson} with each point cloud. As reconstruction requires normal vectors at each point in a cloud, we use the vertex normals derived from the original mesh at each point rather than estimate normals from nearest neighbor tangent planes. The data from the original mesh is then partitioned into $5$ equal subsets, with each subset projected onto a corresponding reconstructed mesh.

\begin{figure}
    \centering
    \includegraphics[width=0.45\textwidth]{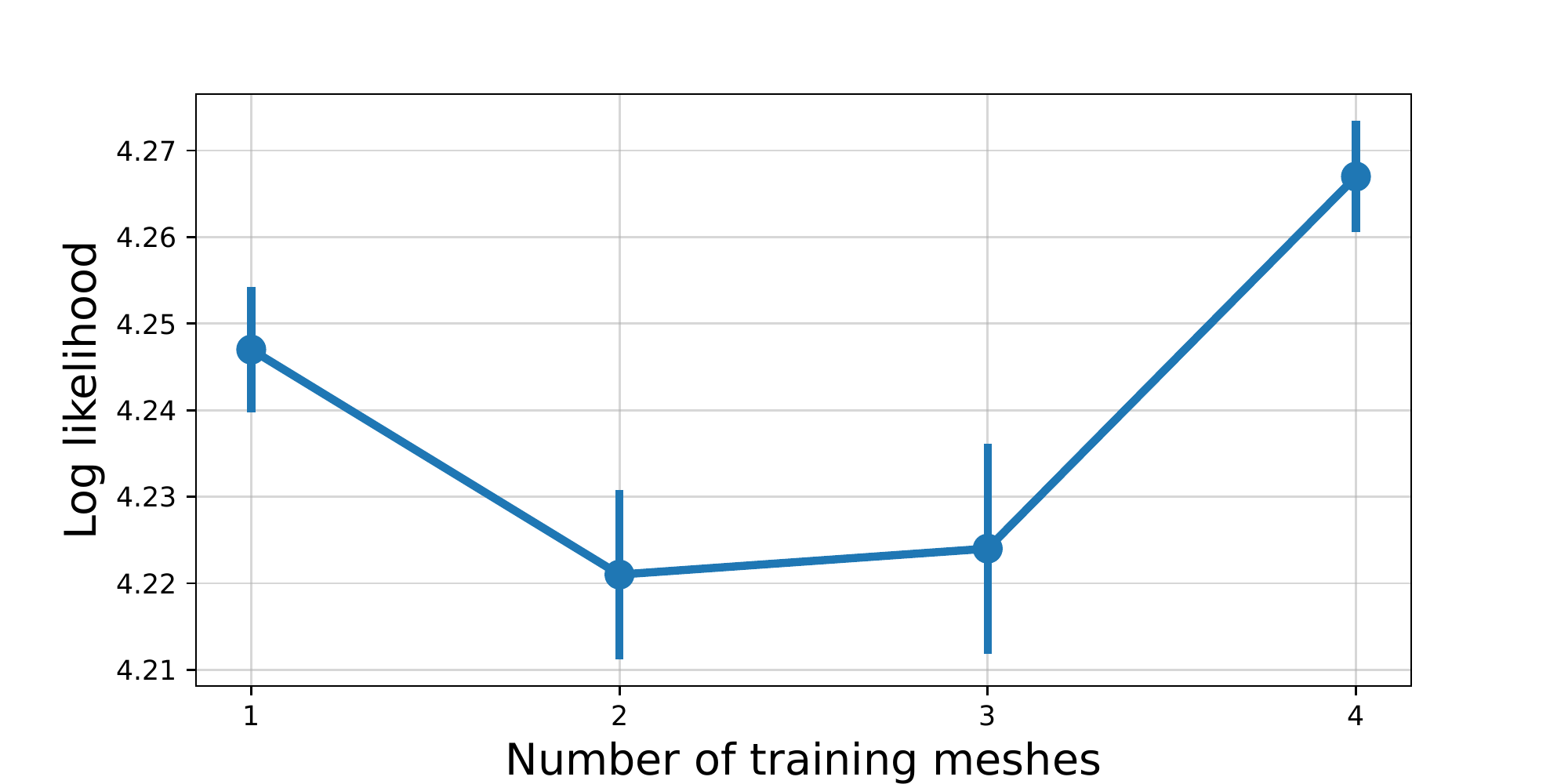}
    \vspace{-0.1in}
    \caption{Performance of generative model on held-out mesh as more meshes and data are added.}
    \label{fig:improvement}
\end{figure}

First, we hold out data from one mesh and run $20$ trials training a CNF with data from $1$, $2$, $3$, and all $4$ of the remaining meshes. Each mesh has $2000$ training data samples; we only use the first $1000$ from each. Log likelihoods of data on the held-out mesh are shown in \cref{fig:improvement}. The CNF used has the same training configuration as in \cref{sec:dist-modeling}.

Second, we train a single conformal generative model that can generate data on each of the five meshes. We use data from each of the five meshes to train the model, again using $1000$ training samples and $1000$ validation samples from each mesh. Generated samples are visualized in \cref{fig:bunnies}. The CNF used has the same training configuration. These results demonstrate that data can be shared between distinct meshes for improved training, and that such models can readily generate data on qualitatively similar but unseen meshes.

\section{Other Related Work}

\textbf{Generative Modeling on Meshes.} The concurrent work of \citet{chen2023riemannian} adopts a flow matching approach to generative modeling on Riemannian manifolds and, as a special case, triangle meshes. In this paradigm, a vector field is trained to match an average (weighted by the data distribution) of conditional vector fields, which are chosen for their relative simplicity. For triangle meshes, the conditional vector fields are constructed using kernel functions encoding approximate spectral distances. Both training and generation with the trained model require solving ODEs directly on the meshes, which can be difficult for complicated geometries.

\textbf{Generative Modeling on Implicit Surfaces.} An alternative approach to generative modeling on complicated manifolds is to a use a learned signed distance function (SDF) \citep{rozen2021moser}, which defines an implicit surface (i.e., the SDF is zero if the point is on the surface).  In general, learning SDFs can be quite complicated, and also lead to uncontrolled approximations of the original manifold.  In contrast, our framework sidesteps this issue by leveraging tools in computational geometery.

\textbf{Conformal Embedding Flows.} Conformal transformations have also been applied to generative modeling problems in high dimensions under the manifold hypothesis \cite{ross2021tractable}. In this case, a (left) invertible map from a low-dimensional latent space to a high-dimensional data space must be learned jointly with a normalizing flow. The embedding is required to be conformal to tractably compute the corresponding change of density; this puts restrictions on the architecture used for manifold learning. Also, as with learning SDFs, manifold learning can be complicated with uncontrolled approximations of the original manifold.


\textbf{Conformal Prediction.} The term ``conformal'' is also used in the context of conformal prediction \citep{shafer2008tutorial}.  Both conformal prediction and conformal geometry use a measure of how conformal a point is under a transformation -- the former in terms of statistical calibration and the latter in terms of geometrical calibration.  Otherwise, the two lines of research are unrelated.


\bibliography{main}
\bibliographystyle{icml2023}

\newpage
\appendix
\onecolumn
\section{Details on Spherical Parameterization}
\label{sec:spherical2}

Here we expand on the spherical parameterization procedure outlined in Section \ref{sec:spherical}.

To compute the flat metric, permitting a planar vertex embedding, we minimize the convex energy in Equation 7 of \citet{springborn2008conformal}, a function of the corresponding log conformal factor. We minimize the energy with second-order optimization (Newton's method), where the gradient of the energy simply measures the defect of the sum of angles around each vertex from $2\pi$ (or $\pi$ for boundary vertices) and the Hessian is the (cotangent) Laplacian of the triangle mesh. When the gradient is $0$, the sum of the angles around each interior vertex is exactly $2\pi$, which is the required curvature of a flat surface. The boundary edge lengths are unchanged by imposing $0$ boundary conditions during optimization.

The domain over which the convex energy is optimized includes log conformal factors corresponding to discrete metrics that violate the triangle inequality, meaning it is possible to find a solution which cannot be used to embed vertices in the plane. \citep{springborn2008conformal} propose flipping edges when such violations are detected, and many follow-up works have investigated this problem further, see \citep{gillespie2021discrete} for a complete discussion and solution. However, we did not encounter such problems in our experiments.

The M\"obius transformations moving the center of the vertex positions can be computed using either \cite{bobenko2016discrete} or \cite{baden2018mobius}; we used the latter, which adds weights to the vertex positions based on corresponding triangle areas from the original mesh.

Denote the discrete metric corresponding to $\tilde{\mathsf{f}}$ by $\tilde{\ell}$. As a byproduct of this procedure, we also obtain the log conformal factor $u$ which establishes discrete conformal equivalence between $\ell$ and $\tilde{\ell}$. We refer to $\tilde{\mathsf{f}}$ as a spherical parametrization of $\mathsf{T}$.

Through spherical parametrization, we establish correspondences between three surfaces: a piecewise linear surface described by $\mathsf{f}$, a piecewise linear surface inscribed within the unit sphere and described by $\tilde{\mathsf{f}}$, and the unit sphere itself. These correspondences are illustrated in \cref{fig:correspondence}. We move between the piecewise linear surfaces via barycentric coordinates within each face (this transformation is also piecewise linear). To move from the surface inscribed within the sphere to the sphere itself, we simply normalize points to have unit norm. To move in the opposite direction (from the sphere to the surface inscribed within the sphere), we compute the intersection of the ray from the origin in the direction of a query point and the inscribed surface. Specifically, if $x \in S^2$ is a query point located in the spherical triangle characterized by $(i, j, k) \in \mathsf{F}$, then the intersection $\alpha x \in \R^3$ (for some scalar $\alpha \in [0, 1]$) with the inscribed surface satisfies:
\begin{equation}
    ((\tilde{\mathsf{f}}_j - \tilde{\mathsf{f}}_i) \times (\tilde{\mathsf{f}}_k - \tilde{\mathsf{f}}_i))^\top (\alpha x - \tilde{\mathsf{f}_i}) = 0,
\end{equation}
where the cross product $(\tilde{\mathsf{f}}_j - \tilde{\mathsf{f}}_i) \times (\tilde{\mathsf{f}}_k - \tilde{\mathsf{f}}_i)$ is perpendicular to the inscribed Euclidean triangle characterized by $(i, j, k)$. Rearranging terms, we have:
\begin{equation}\label{eqn:scale-factor}
    \alpha = \frac{((\tilde{\mathsf{f}}_j - \tilde{\mathsf{f}}_i) \times (\tilde{\mathsf{f}}_k - \tilde{\mathsf{f}}_i))^\top \tilde{\mathsf{f}}_i}{((\tilde{\mathsf{f}}_j - \tilde{\mathsf{f}}_i) \times (\tilde{\mathsf{f}}_k - \tilde{\mathsf{f}}_i))^\top x} = \frac{( \tilde{\mathsf{f}}_j \times \tilde{\mathsf{f}}_k )^\top \tilde{\mathsf{f}}_i}{((\tilde{\mathsf{f}}_j - \tilde{\mathsf{f}}_i) \times (\tilde{\mathsf{f}}_k - \tilde{\mathsf{f}}_i))^\top x}.
\end{equation}
This latter correspondence is illustrated in \cref{fig:correspondence}.

With these correspondences established, we denote the two piecewise linear surfaces as $\mathsf{M} \subset \R^d$ and $\mathsf{M}_{\mathrm{inscr}} \subset \R^3$, respectively. We also generate an interpolation of the log conformal factor $u$ as a function on the sphere, $\bar{u}: S^2 \to \R$. To compute the interpolated value $\bar{u}(x)$ at a query point $x \in S^2$, we first project $x$ onto $\mathsf{M}_{\mathrm{inscr}}$ using \cref{eqn:scale-factor}; we then interpolate the values of $u$ at the vertices of the intersected triangle using the corresponding barycentric coordinates as interpolation weights.

\section{Details on Base Generative Models}
\label{sec:base2}

\subsection{Riemannian Continuous Normalizing Flows}
\label{sec:CNF}

In the sections that follow, for convenience we adopt the convention of diffusion and score-based generative models \cite{sohl2015deep,song2020score}, for which the \textit{forward} direction of a generative model maps a data distribution to a noise distribution.

In the Euclidean setting, a \textit{continuous normalizing flow} (CNF) \cite{chen2018neural} is characterized by a parametric time-varying vector field $f: \R^d \times [0, 1] \to \R^d$. For an initial probability density $\rho_0: \R^d \to \R_+$, the time-varying density $\rho: \R^d \times [0, 1] \to \R_+$ solving (with initial condition $\rho_0$) the probability mass continuity equation:
\begin{equation}\label{eqn:continuity-eq}
    \frac{\partial \rho}{\partial t} + \mathrm{div}(\rho \cdot f) = 0,
\end{equation}
describes how the probability density of a random initial condition (distributed according to $\rho_0$) evolves subject to the vector field $f$. The product $\rho \cdot f$ is the \textit{probability mass flux}, and its divergence can be expanded as:
\begin{equation}
    \mathrm{div}(\rho \cdot f) = \nabla \rho^\top f + \rho \cdot \mathrm{div}f.
\end{equation}
This form allows us to express the rate of change of (log) probability density along deterministic trajectories governed by the vector field $f$. That is, for $x \in \R^d$ and a trajectory $\gamma: [0, 1] \to \R^d$ satisfying:
\begin{equation}\label{eqn:state-evolution}
    \frac{\mathrm{d}\gamma}{\mathrm{d}t} = f(\gamma(t), t),
\end{equation}
for all $t \in (0, 1)$, we also have (from \citet{chen2018neural}):
\begin{equation}\label{eqn:log-prob-evolution}
    \frac{\mathrm{d}}{\mathrm{d}t}\log{\rho(\gamma(t), t)} = -\mathrm{div}(\rho \cdot f)(\gamma(t), t).
\end{equation}
This ordinary differential equations represented by \cref{eqn:state-evolution} and \cref{eqn:log-prob-evolution} can be solved simultaneously. With this augmented system, we can map a data distribution to a noise distribution (solve forward) or a noise distribution to a data distribution (solve backward). Training the model requires querying log likelihoods of data; to compute likelihoods, data are propagated forward under \cref{eqn:state-evolution}, the log probability densities of the corresponding terminal states (under a chosen noise distribution) are computed, and the terminal states and log probability densities are propagated backward under both \cref{eqn:state-evolution} and \cref{eqn:log-prob-evolution}.

\textbf{From Euclidean to Spherical Gradients.}
In the spherical setting, vector fields specify tangent vectors. That is, at a point $x \in S^2$, a vector field specifies a tangent vector $v \in \R^3$ such that $x^\top v = 0$. We can compute differential operators like $\nabla$ and $\mathrm{div}$ and solve ordinary differential equations by extending functions defined on the sphere to functions defined on all of $\R^3 \setminus \{ 0 \}$ as in \citet{rozen2021moser}. Specifically, for tangent vector field $f: S^2 \times [0, 1] \to \R^3$ and probability density $\rho: S^2 \times [0, 1] \to \R_+$, we define extensions:
\begin{align}
    \tilde{f}(x, t) &= f\left(x / \| x \|_2, t\right), & \tilde{\rho}(x, t) &= \rho\left(x / \| x \|_2, t\right),
\end{align}
for all nonzero $x \in \R^3$ and times $t \in [0, 1]$. The spherical gradient of $f$ is the Euclidean gradient of $\tilde{f}$. The spherical divergences of $f$ and $\rho \cdot f$ are the respective Euclidean divergences of $\tilde{f}$ and $\tilde{\rho} \cdot \tilde{f}$. As in \cite{rozen2021moser}, we can solve \cref{eqn:state-evolution} by substituting $f$ with $\tilde{f}$ and deploying any differentiable (adaptive-step) Euclidean solver. Alternatively, we can solve \cref{eqn:state-evolution} using charts on the sphere \cite{lou2020neural}, which locally represent the sphere and the vector field in Euclidean space (for example, by using spherical exponential and log maps). While this latter approach has the advantage of constructively restricting the solution of \cref{eqn:state-evolution} to the sphere, in our experiments the former approach maintained proximity to the sphere within tight tolerances.

\subsection{Moser Flows}
\label{sec:moser}

\citet{rozen2021moser} bypass the need to solve ordinary differential equations to query log likelihoods by modeling the probability mass flux directly (as opposed to the vector field $f$) and restricting the class of vector fields used to generate data samples. For a data density $\mu: S^2 \to \R_+$ and a noise density $\nu: S^2 \to \R_+$, the probability mass flux is modeled as a parametric time-invariant tangent vector field $F: S^2 \to \R^3$ for which:
\begin{equation}\label{eqn:moser}
    \mu(x) - \nu(x) = -\mathrm{div}F(x),
\end{equation}
for all $x \in S^2$. That is, when $\nu$ is selected as well as parameters for $F$, the modeled data density is given by \cref{eqn:moser}. To generate data samples from noise, the vector field:
\begin{equation}
    f(x, t) = \frac{F(x)}{\nu(x) + (1 - t) \cdot \mathrm{div}F(x)},
\end{equation}
is used, and the probability density solving the corresponding continuity equation is:
\begin{equation}
    \rho(x, t) = (1 - t)\mu(x) + t\nu(x),
\end{equation}
for all $x \in S^2$ and $t \in [0, 1]$. Note that this probability density interpolates linearly (in time) between $\mu$ and $\nu$.

To ensure that the flux model corresponds to a valid probability density, the divergence of the flux field must integrate to $0$ and be greater than $-\nu$ everywhere. The first requirement follows from the divergence theorem, replacing the integral of divergence with a boundary integral (where the boundary is empty). The second requirement must be enforced via an integral constraint, approximated with Monte Carlo integration (using a uniform samples from the original mesh).


\end{document}